\documentclass{article} 
\usepackage{iclr2018_conference,times}
\usepackage{hyperref}
\usepackage{url}
\usepackage{graphicx}
\usepackage{epstopdf}
\usepackage{indentfirst}
\usepackage{amsmath}
\usepackage{enumerate}
\usepackage{multirow}
\usepackage{array}
\usepackage{bbding}
\usepackage{booktabs}
\usepackage{multirow}
\usepackage{multicol}
\usepackage{algorithm}
\usepackage{algorithmic}
\usepackage{cleveref}
\usepackage{color}
\usepackage{mathrsfs}
\usepackage{amsfonts}
\usepackage{bm}
\usepackage{textcomp}

\iclrfinalcopy

\title{Training and Inference with Integers in Deep Neural Networks}

\author{Shuang Wu$^{1}$, Guoqi Li$^{1}$, Feng Chen$^{2}$, Luping Shi$^{1}$  \\ $^{1}$Department of Precision Instrument\\
$^{2}$Department of Automation \\
Center for Brain Inspired Computing Research\\
Beijing Innovation Center for Future Chip\\
Tsinghua University \\
\texttt{\{lpshi,chenfeng\}@mail.tsinghua.edu.cn} \\
}

\begin{document}

\maketitle

\begin{abstract}

Researches on deep neural networks with discrete parameters and their deployment in embedded systems have been active and promising topics.
Although previous works have successfully reduced precision in inference, transferring both training and inference processes to low-bitwidth integers has not been demonstrated simultaneously.
In this work, we develop a new method termed as ``WAGE" to discretize both training and inference, where weights (W), activations (A), gradients (G) and errors (E) among layers are shifted and linearly constrained to low-bitwidth integers.
To perform pure discrete dataflow for fixed-point devices, we further replace batch normalization by a constant scaling layer and simplify other components that are arduous for integer implementation.
Improved accuracies can be obtained on multiple datasets, which indicates that WAGE somehow acts as a type of regularization.
Empirically, we demonstrate the potential to deploy training in hardware systems such as integer-based deep learning accelerators and neuromorphic chips with comparable accuracy and higher energy efficiency, which is crucial to future AI applications in variable scenarios with transfer and continual learning demands.
\end{abstract}

\section{Introduction}

Recently deep neural networks (DNNs) are being widely used for numerous AI applications \citep{krizhevsky2012imagenet,hinton2012deep,silver2016mastering}. Depending on the massive tunable parameters, DNNs are considered to have powerful multi-level feature extraction and representation abilities. However, training DNNs needs energy-intensive devices such as GPU and CPU with high precision (float32) processing units and abundant memory, which has greatly challenged their extensive applications for portable devices. In addition, a state-of-art network often has far more weights and effective capacity to shatter all training samples \citep{zhang2016understanding}, leading to overfitting easily.

As a result, there is much interest in reducing the size of network during inference \citep{hubara2016binarized,rastegari2016xnor,li2016ternary}, as well as dedicated hardware for commercial solutions \citep{jouppi2017datacenter,chen2017eyeriss,shi2015development}. Due to the accumulation in stochastic gradient descent (SGD) optimization, the precision demand for training is usually higher than inference \citep{hubara2016binarized,li2017training}. Therefore, most of the existing techniques only focus on the deployment of a well-trained compressed network, while still keeping high precision and computational complexity during training. In this work, we address this problem as how to process both training and inference with low-bitwidth integers, which is essential for implementing DNNs in dedicated hardware. To this end, two fundamental issues are addressed for discretely training DNNs: i) how to quantize all the operands and operations, and ii) how many bits or states are needed for SGD computation and accumulation.

With respect to the issues, we propose a framework termed as ``WAGE" that constrains weights (W), activations (A), gradients (G) and errors (E) among all layers to low-bitwidth integers in both training and inference. Firstly, for operands, linear mapping and orientation-preserved shifting are applied to achieve ternary weights, 8-bit integers for activations and gradients accumulation. Secondly, for operations, batch normalization \citep{ioffe2015batch} is replaced by a constant scaling factor. Other techniques for fine-tuning such as SGD optimizer with momentum and L2 regularization are simplified or abandoned with little performance degradation. Considering the overall bidirectional propagation, we completely streamline inference into accumulate-compare cycles and training into low-bitwidth multiply-accumulate (MAC) cycles with alignment operations, respectively.

We heuristically explore the bitwidth requirements of integers for error computation and gradient accumulation, which have rarely been discussed in previous works. Experiments indicate that it is the relative values (orientations) rather than absolute values (orders of magnitude) in error that guides previous layers to converge. Moreover, small values have negligible effects on previous orientations though propagated layer by layer, which can be partially discarded in quantization. We leverage these phenomena and use an orientation-preserved shifting operation to constrain errors. As for the gradient accumulation, though weights are quantized to ternary values in inference, a relatively higher bitwidth is indispensable to store and accumulate gradient updates.

The proposed framework is evaluated on MNIST, CIFAR10, SVHN, ImageNet datasets. Comparing to those who only discretize weights and activations at inference time, it has comparable accuracy and can further alleviate overfitting, indicating some type of regularization. WAGE produces pure bidirectional low-precision integer dataflow for DNNs, which can be applied for training and inference in dedicated hardware neatly. We publish the code on GitHub\footnote{https://github.com/boluoweifenda/WAGE}.

\section{Related Work}

We mainly focus on reducing precision of operands and operations in both training and inference. Orthogonal and complementary techniques for reducing complexity like network compression, pruning \citep{han2015deep,zhou2017incremental} and compact architectures \citep{howard2017mobilenets} are impressively efficient but outside the scope this paper.

\textbf{Weight and activation} \citet{courbariaux2015binaryconnect,hubara2016binarized} propose methods to train DNNs with binary weights (BC) and activations (BNN) successively. They add noises to weights and activations as a form of regularization but real-valued gradients are accumulated in real-valued variables, suggesting that high precision accumulation is likely required for SGD optimization. XNOR-Net \citep{rastegari2016xnor} has a filter-wise scaling factor for weights to improve the performance. Convolutions in XNOR-Net can be implemented efficiently using XNOR logical units and bit-count operations. However, these floating-point factors are calculated simultaneously during training, which generally aggravates the training effort. In TWN \citep{li2016ternary} and TTQ \citep{zhu2016trained} two symmetric thresholds are introduced to constrain the weights to be ternary-valued: $\{+1, 0,-1\}$. They claimed a tradeoff between model complexity and expressive ability.

\textbf{Gradient computation and accumulation} DoReFa-Net \citep{zhou2016dorefa} quantizes gradients to low-bitwidth floating-point numbers with discrete states in the backward pass. TernGrad \citep{wen2017terngrad} quantizes gradient updates to ternary values to reduce the overhead of gradient synchronization in distributed training. Nevertheless, weights in DoReFa-Net and TernGrad are stored and updated with float32 during training like previous works. Besides, the quantization of batch normalization and its derivative is ignored. Thus, the overall computation graph for the training process is still presented with float32 and more complex with external quantization. Generally, it is difficult to apply DoReFa-Net training in an integer-based hardware directly, but it shows potential for exploring high-dimensional discrete spaces with discrete gradient descent directions.

\section{WAGE Quantization}

\begin{figure}[h]
\begin{center}
\includegraphics[width=5.5in]{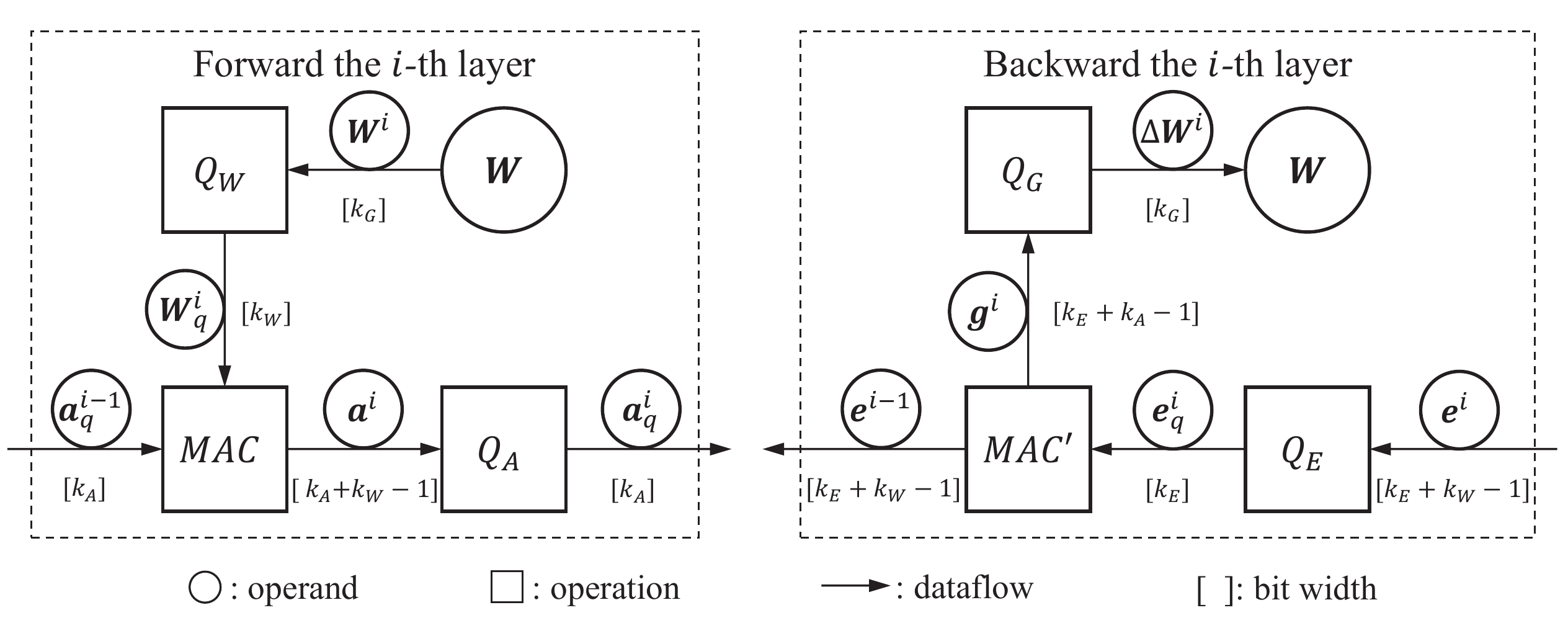}
\end{center}
\caption{Four operators $Q_W(\cdot)$, $Q_A(\cdot)$, $Q_G(\cdot)$, $Q_E(\cdot)$ added in WAGE computation dataflow to reduce precision, bitwidth of signed integers are below or on the right of arrows, activations are included in MAC for concision.}
\label{fig1}
\end{figure}

The main idea of WAGE quantization is to constrain four operands to low-bitwidth integers: weight $\bm{W}$ and activation $\bm{a}$ in inference, error $\bm{e}$ and gradient $\bm{g}$ in backpropagation training, see Figure~\ref{fig1}. We extend the original definition of errors to multi-layer: error $\bm{e}$ is the gradient of activation $\bm{a}$ for the perspective of each convolution or fully-connected layer, while gradient $\bm{g}$ particularly refers to the gradient accumulation of weight $\bm{W}$. Considering the $i$-th layer of a feed-forward network, we have:
\begin{equation}
  \label{eqn1}
  \bm{e}^i=\frac{\partial{\mathcal{L}}}{\partial{\bm{a}^i}},\
  \bm{g}^i=\frac{\partial{\mathcal{L}}}{\partial{\bm{W}^i}}
\end{equation}
where $\mathcal{L}$ is the loss function. We separate these two terms that are mixed up in most existing schemes. The gradient of weight $\bm{g}$ and the gradient of activation $\bm{e}$ flow to different paths in each layer, which is a fork both in inference and in backward training and generally acts as node of MAC operations.

For the forward propagation in the $i$-th layer, assuming that weights are stored and accumulated with $k_G$-bit integers, then numerous works strive for a better quantization function $Q_W(\cdot)$ that maps higher precision weights to their $k_W$-bit reflections, for example, $[-0.9, 0.1, 0.7]$ to $[-1, 0, 1]$. Although weights are accumulated with high precision like float32, the deployment of the reflections in dedicated hardware are much more memory efficient after training. Activations are quantized with function $Q_A(\cdot)$ to $k_A$ bits to align the increased bitwidth caused by MACs. Weights and activations are discretized to even binary values in previous works, then MACs degrade into logical and bit-count operations that are extremely efficient \citep{rastegari2016xnor}.

For the backward propagation in the $i$-th layer, the gradients of activations and weights are calculated by the derivatives of MACs that are generally considered to be in 16-bit floating-point precision at least. As illustrated in Figure~\ref{fig1}, the MACs between $k_A$-bit inputs and $k_W$-bit weights will increase the bitwidth of outputs to $[k_A+k_W-1]$ in signed integer representation, and the similar broadening happens to errors $\bm{e}$ as well. In consideration of training with only low-bitwidth integers, we propose additional functions $Q_E(\cdot)$ and $Q_G(\cdot)$ to constrain bitwidth of $\bm{e}$ and $\bm{g}$ to $k_E$ bits and $k_G$ bits, respectively. In general, where there is a MAC operation, there are quantization operators named $Q_W(\cdot)$, $Q_A(\cdot)$, $Q_G(\cdot)$ and $Q_E(\cdot)$ in inference and backpropagation.

\subsection{Shift-based Linear Mapping and Stochastic Rounding}
In WAGE quantization, we adopt a linear mapping with $k$-bit integers for simplicity, where continuous and unbounded values are discretized with uniform distance $\sigma$:
\begin{equation}
  \sigma(k)=2^{1-k},k\in\mathbb{N_+}
\end{equation}
Then the basic quantization function that converts a floating-point number $x$ to its $k$-bitwidth signed integer representation can be formulated as:
\begin{equation}
  \label{eqn2}
  Q(x,k)=Clip\left\{\  \sigma(k) \cdot round\left[\frac{x}{\sigma(k)}\right], -1+\sigma(k), 1-\sigma(k) \ \right\}
\end{equation}
where $round$ approximates continuous values to their nearest discrete states. $Clip$ is the saturation function that clips unbounded values to $[-1+\sigma,1-\sigma]$, where the negative maximum value $-1$ is removed to maintain symmetry. For example, $Q(x,2)$ quantizes $\{-1,0.2,0.6\}$ to $\{-0.5,0,0.5\}$. Equation~\ref{eqn2} is merely used for simulation in floating-point hardware like GPU, whereas in a fixed-point device, quantization and saturation is satisfied automatically.

\begin{figure}[h]
\begin{center}
\includegraphics[width=5.5in]{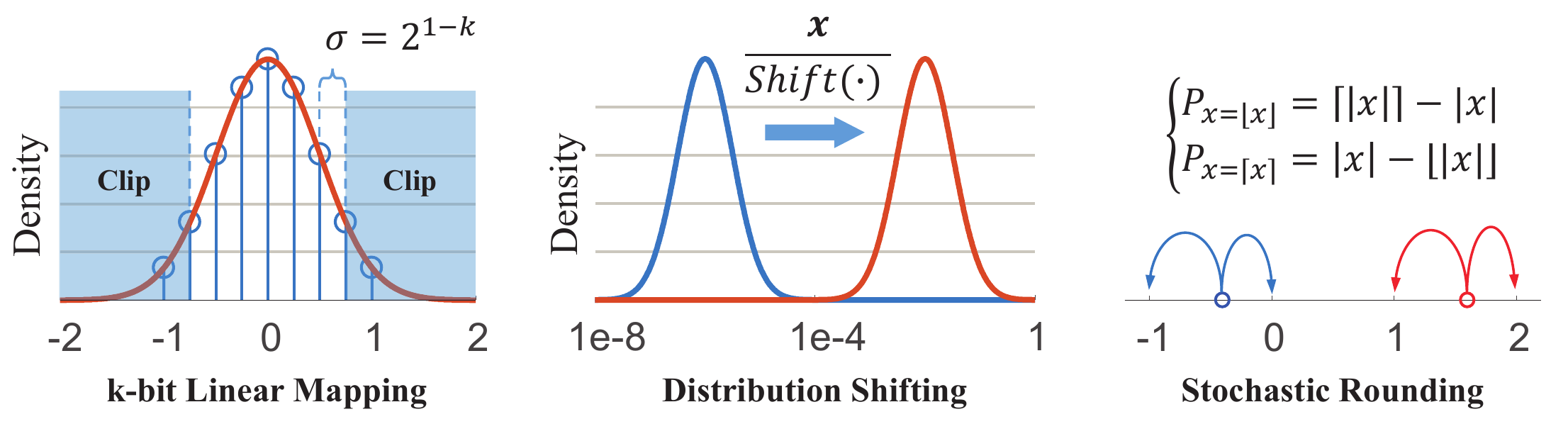}
\end{center}
\caption{Quantization methods used in WAGE. The notation $P$, $\bm{x}$, $\lfloor \cdot \rfloor$ and $\lceil \cdot \rceil$ denotes probability, vector, $floor$ and $ceil$, respectively. $Shift(\cdot)$ refers to Equation~\ref{eqn4} with a certain argument.}
\label{fig4}
\end{figure}

Before applying linear mapping in some operands (e.g., error), we introduce an additional monolithic scaling factor for shifting values distribution to an appropriate order of magnitude, otherwise values will be all saturated or cleared by Equation~\ref{eqn2}. The scaling factor is calculated by $Shift$ function and then divided in later steps:
\begin{equation}
  \label{eqn4}
  Shift(x)=2^{round(\log_2{x})}
\end{equation}
Finally, we propose stochastic rounding to substitute small and real-valued updates for gradient accumulation in training. Section~\ref{gradient} will detail the implementation of operator $Q_G(\cdot)$, where high bitwidth gradients are constrained to $k_G$-bit integers stochastically by a 16-bit random number generator. Figure~\ref{fig4} summarizes quantization methods used in WAGE.

\subsection{Weight Initialization}
In previous works, weights are binarized directly by $sgn$ function or ternarized by threshold parameters calculated during training. However, BNN fails to converge without batch normalization because weight values $\pm 1$ are rather big for a typical DNN. Batch normalization not only efficiently avoids the problem of exploding and vanishing gradients, but also alleviates the demand for proper initialization. However, normalizing outputs for each layer and computing their gradients are quite complex without floating point unit (FPU). Besides, the moving averages of batch outputs occupy external memory. BNN shows a shift-based variation of batch normalization but it is hard to transform all of the elements to the fixed-point representations. As a result, weights should be cautiously initialized in this work where batch normalization is simplified to a constant scaling layer. A modified initialization method based on MSRA \citep{he2015delving} can be formulated as:
\begin{equation}
\label{eqn5}
  \bm{W} \sim U(-L,+L), L=max\{\sqrt{6/n_{in}},L_{min}\},L_{min}=\beta\sigma
\end{equation}
where $n_{in}$ is the layer fan-in number, and the original limit $\sqrt{6/n_{in}}$ in MSRA is calculated to keep same variance between inputs and outputs of the same layer theoretically. The additional limit $L_{min}$ is a minimum value that the uniform distribution $U$ should reach, and $\beta$ is a constant greater than $1$ to create overlaps between minimum step size $\sigma$ and maximum value $L$. In case of $k_W$-bit linear mapping, if weights $\bm{W}$ are quantized directly with original limits, we will get all-zero tensors when bitwidth $k_W$ is small enough, e.g., $4$, or fan-in $n_{in}$ is wide enough, where initialized weights may never reach the minimum step $\sigma$ presented by fixed-point integers. So $L_{min}$ ensures that weights can go beyond $\sigma$ and quantized to non-zero values after $Q_W(\cdot)$ when initialized randomly.

\subsection{Quantization Details}
\subsubsection{Weight $Q_W(\cdot)$}
The modified initialization in Equation~\ref{eqn5} will amplify weights holistically and guarantee their proper distribution, then $\bm{W}$ is quantized directly with Equation~\ref{eqn2}:
\begin{equation}
\label{eqn8}
  \bm{W}_q=Q_W(\bm{W})=Q(\bm{W},k_W)
\end{equation}
It should be noted that the variance of weights is scaled compared to the original limit, which will cause exploding of network's outputs. To alleviate the amplification effect, XNOR-Net proposed a filter-wise scaling factor calculated continuously with full precision. In consideration of integer implementation, we introduce a \emph{layer-wise shift-based} scaling factor $\alpha$ to attenuate the amplification effect:
\begin{equation}
\label{eqn6}
  \alpha=max \{ Shift(L_{min}/L),1 \}
\end{equation}
where $\alpha$ is a pre-defined constant for each layer determined by the network structure. The modified initialization and attenuation factor $\alpha$ together approximates floating-point weights to their integer representations, except that $\alpha$ takes effect after activations to maintain precision of weights presented by $k_W$-bit integers.

\subsubsection{Activation $Q_A(\cdot)$}
As stated above, the bitwidth of operands increases after MACs. Then a typical CNN is usually followed with pooling, normalization and activation. Average pooling is avoided because mean operations will increase precision demand. Besides, we hypothesize that batch outputs of each hidden layer approximately have zero-mean, then batch normalization degenerates into to a scaling layer where trainable and batch-calculated scaling parameters are replaced by $\alpha$ mentioned in Equation~\ref{eqn6}. If activations are presented in $k_A$ bits, the overall quantization of activations can be formulated as:
\begin{equation}
\label{eqn9}
  \bm{a}_q=Q_A(\bm{a})=Q(\bm{a}/\alpha,k_A)
\end{equation}

\subsubsection{Error $Q_E(\cdot)$}
Errors $\bm{e}$ are calculated layer by layer using the chain rule during training. Although the computation graph of backpropagation is similar to the inference, the inputs are the gradients of $\mathcal{L}$, which are relatively small compared to actual inputs for networks. More importantly, the errors are unbounded and might have significantly larger ranges than that of activations, e.g., $[10^{-9}, 10^{-4}]$. DoReFa-Net first applies an affine transform on $\bm{e}$ to map them into $[-1, 1]$, and then inverts the transform after quantization. Thus, the quantized $\bm{e}$ are still presented as float32 numbers with discrete states and mostly small values.

However, experiments uncover that it is the orientations rather than orders of magnitude in errors that guides previous layers to converge, then the inverse transformation after quantization in DoReFa-Net is no longer needed. The orientation-only preservation prompts us to propagate errors with integer thoroughly, where error distribution is firstly scaled into $[-\sqrt{2},+\sqrt{2}]$ by dividing a shift factor as shown in Figure~\ref{fig4} and then quantized by $Q(\bm{e},k_E)$:
\begin{equation}
\label{eqn10}
  \bm{e}_q=Q_E(\bm{e})=Q(\bm{e}/Shift(max\{|\bm{e}|\}),k_E)
\end{equation}
where $max\{|\bm{e}|\}$ extracts the layer-wise maximum absolute value among all elements in error $\bm{e}$, multi-channel for convolution and multi-sample for batch training. The quantization of error discards large proportion of values smaller than $\sigma$, we will discuss the influence on accuracy later.

\subsubsection{Gradient $Q_G(\cdot)$}
\label{gradient}
Since we only preserve relative values of error after shifting, the gradient updates $\bm{g}$ derived from MACs between backward errors $\bm{e}$ and forward activations $\bm{a}$ are shifted consequently. We first rescale gradients $\bm{g}$ with another scaling factor and then bring in shift-based learning rate $\eta$:
\begin{equation}
\label{eqn11}
  \bm{g}_s=\eta \cdot \bm{g}/Shift(max\{|\bm{g}|\})
\end{equation}
where $\eta$ is an integer power of 2. The shifted gradients $\bm{g}_s$ represent for minimum step numbers and directions for updating weights. If weights are stored with $k_G$-bit numbers, the minimum step of modification will be $\pm 1$ for integers and $\pm \sigma(k_G)$ for floating-point values, respectively. The implement of learning rate $\eta$ here is quite different from that in a vanilla DNN based on float32. In WAGE, there only remain directions for weights to change and the step sizes are integer multiples of minimum step $\sigma$. Shifted gradients $\bm{g}_s$ may get greater than $1$ if $\eta$ is $2$ or bigger to accelerate training at the beginning, or smaller than $0.5$ during latter half of training when learning rate decay is usually applied. As illustrated in Figure~\ref{fig4}, to substitute accumulation of small gradients in latter case, we separate $\bm{g}_s$ into integer parts and decimal parts, then use a 16-bit random number generator to constrain high bitwidth $\bm{g}_s$ to $k_G$-bit integers stochastically:
\begin{equation}
\label{eqn12}
  \Delta{\bm{W}}=Q_G(\bm{g})= \sigma(k_G) \cdot sgn(\bm{g}_s) \cdot
  \Big{\{} \
  \lfloor|\bm{g}_s|\rfloor  +
  Bernoulli(\
  |\bm{g}_s|-\lfloor|\bm{g}_s|\rfloor
  \ ) \ \Big{\}}
\end{equation}
where $Bernoulli$ \citep{zhou2016dorefa} stochastically samples decimal parts to either $0$ or $1$. With proper setting of $k_G$, quantization of gradients will restrict the minimum step size, which may avoid local minimum and overfitting. Furthermore, the gradients will be ternary values when $\eta$ is not greater than $1$, which reduces communication costs for distributed training \citep{wen2017terngrad}. At last, weights $\bm{W}$ might exceed the range $[-1+\sigma,1-\sigma]$ presented by $k_G$-bit integers after updating with discrete increments $\Delta{\bm{W}}$. So $Clip$ function is indispensable to saturate and make sure there are only $2^{k_G-1}-1$ states for weights accumulation. In case of the $t$-th iteration, we have:
\begin{equation}
\label{eqn13}
  \bm{W}_{t+1}  = Clip\left\{
  \bm{W}_t - \Delta{\bm{W}_t},
  -1+\sigma(k_G),
  1-\sigma(k_G)\right\}
\end{equation}

\subsection{Miscellaneous}
From the above, we have illustrated our quantization methods for weights, activations, gradients and errors. See Algorithm~\ref{algorithm1} for the detailed computation graph. There remain some issues to specify in an overall training process with only integers.

Gradient descent optimizer like Momentum, RMSProp and Adam contains at least one copy of gradient updates $\Delta{\bm{W}}$ or their moving average, doubling memory consumption for weights during training, which is partially equivalent to use bigger $k_G$. Since the weight updates $\Delta{\bm{W}}$ are quantized to integer multiple of $\sigma$ and scaled by $\eta$, we adopt pure mini-batch SGD without any form of momentum or adaptive learning rate to show the potential of reducing storage demands.

Although L2 regularization works quite well for many large-scale DNNs where overfitting occurs commonly, WAGE removes small values in Equation~\ref{eqn2} and introduces randomness in Equation~\ref{eqn12}, acting as certain types of regularization and can get comparable accuracy in later experiments. Thus, we remain L2 weight decay and dropout as supplementary regularization methods.

The \emph{Softmax} layer and cross-entropy criterion are widely adopted in classification tasks but the calculation of $e^x$ can hardly be applied in low-bitwidth linear mapping occasions. For tasks with small number of categories, we avoid \emph{Softmax} layer and apply mean-square-error criterion but omit mean operation to form a sum-square-error (SSE) criterion since shifted errors will get the same values in Equation~\ref{eqn10}.

\section{Experiments}
In this section, we set W-A-G-E bits to 2-8-8-8 as default for all layers in a CNN or MLP. The bitwidth $k_W$ is $2$ for ternary weights, which implies that there are no multiplications during inference. Constant parameter $\beta$ is $1.5$ to make equal probabilities for ternary weights when initialized randomly. Activations and errors should be of the same bitwidth since computation graph of backpropagation is similar to inference and might be applied in the same partition of hardware or memristor array \citep{sheridan2017sparse}. Although XNOR-Net achieves 1-bit activations, reducing errors to 4 or less bits dramatically degenerates accuracies in our tests, so the bitwidth $k_A$ and $k_E$ are increased to 8 simultaneously. Weights are stored with 8-bit integers during training and ternarized by two constant symmetrical thresholds during inference.
We first build the computation graph for a vanilla network, then insert quantization nodes in forward propagation and override gradients in backward propagation for each layer on Tensorflow \citep{abadi2016tensorflow}.
Our method is evaluated on MNIST, SVHN, CIFAR10 and ILSVRC12 \citep{russakovsky2015imagenet} and Table~\ref{tab1} shows the comparison results.

\subsection{Implement Details}

\textbf{MNIST:} A variation of LeNet-5 \citep{lecun1998gradient} with 32C5-MP2-64C5-MP2-512FC-10SSE is adopted. The input grayscale images are regarded as activations and quantized by Equation~\ref{eqn9} where $\alpha$ equals to 1. The learning rate $\eta$ in WAGE remains as $1$ for the whole 100 epochs. We report average accuracy of 10 runs on the test set.

\textbf{SVHN \& CIFAR10:} We use a VGG-like network \citep{simonyan2014very} with 2$\times$(128C3)-MP2-2$\times$(256C3)-MP2-2$\times$(512C3)-MP2-1024FC-10SSE. For CIFAR10 dataset, we follow the data augmentation in \citet{lee2015deeply} for training: 4 pixels are padded on each side, and a 32$\times$32 patch is randomly cropped from the padded image or its horizontal flip. For testing, only single view of the original 32$\times$32 image is evaluated. The model is trained with mini-batch size of 128 and totally 300 epochs. Learning rate $\eta$ is set to 8 and divided by 8 at epoch 200 and epoch 250. The original images are scaled and biased to the range of $[-1,+1]$ for 8-bit integer activation representation. As for SVHN dataset, we leave out randomly flip augmentation and reduce training epochs to 40 since it is a rather big dataset. The error rate is evaluated in the same way as MNIST.

\textbf{ImageNet:} WAGE framework is evaluated on ILSVRC12 dataset with AlexNet \citep{krizhevsky2012imagenet} model but removes dropout and local response normalization layers. Images are firstly resized to 256$\times$256 then randomly cropped to 224$\times$224 and horizontally flipped, followed by bias subtraction as CIFAR10. For testing, the single center crop in validation set is evaluated. Since ImageNet task is much difficult than CIFAR10 and has 1000 categories, it is hard to converge when applying SSE or hinge loss criterion in WAGE, so we add \emph{Softmax} and remove quantizations in the last layer for fear of severe accuracy drop \citep{tang2017train}. The model is trained with mini-batch size of 256 and totally 70 epochs. Learning rate $\eta$ is set to 4 and divided by 8 at epoch 60 and epoch 65.

\renewcommand\arraystretch{1.2}
\begin{table}[h]
\caption{Test or validation error rates (\%) in previous works and WAGE on multiple datasets. Opt denotes gradient descent optimizer and withM means SGD with momentum, BN represents for batch normalization and 32 bits refers to float32, ImageNet top-k format: top1/top5.}
\begin{center}
\begin{tabular}{ p{1.3cm}<{\centering}
                 p{0.5cm}<{\centering}
                 p{0.5cm}<{\centering}
                 p{0.65cm}<{\centering}
                 p{0.65cm}<{\centering}
                 p{0.80cm}<{\centering}
                 p{0.4cm}<{\centering}
                 p{1.0cm}<{\centering}
                 p{0.8cm}<{\centering}
                 p{1.2cm}<{\centering}
                 p{1.4cm}<{\centering}
                 }
\toprule
Method & $k_W$ & $k_A$  & $k_G$  & $k_E$  & Opt & BN   & MNIST & SVHN & CIFAR10 & ImageNet  \\
\hline
BC      & 1 & 32 & 32 & 32 & Adam  & \Checkmark & 1.29  & 2.30 & 9.90 & - \\
BNN     & 1 & 1  & 32 & 32 & Adam  & \Checkmark & 0.96  & 2.53 &10.15 & - \\
BWN\footnotemark[1]
        & 1 & 32 & 32 & 32 & withM & \Checkmark & -     & -    & -    & 43.2/20.6 \\
XNOR    & 1 & 1  & 32 & 32 & Adam  & \Checkmark & -     & -    & -    & 55.8/30.8 \\
TWN     & 2 & 32 & 32 & 32 & withM & \Checkmark & 0.65  & -    & 7.44 & 34.7/13.8 \\
TTQ     & 2 & 32 & 32 & 32 & Adam  & \Checkmark & -     & -    & 6.44 & 42.5/20.3 \\
DoReFa\footnotemark[2]  & 8 & 8 & 32 & 8 & Adam& \Checkmark& -    & 2.30 & -    & 47.0/\ \ \ -\ \ \ \ \quad \\
TernGrad\footnotemark[3] & 32 & 32 & 2 & 32 & Adam & \Checkmark & - & - & 14.36 & 42.4/19.5  \\
WAGE    & 2 & 8  & 8 & 8 & SGD & \XSolidBrush & \textbf{0.40}  & \textbf{1.92} & 6.78    & 51.6/27.8 \\
\bottomrule
\end{tabular}
\end{center}
\label{tab1}
\end{table}
\footnotetext[1]{BWN is the counterpart of XNOR that only quantizes weights}
\footnotetext[2]{They use floating-point presentation for errors}
\footnotetext[3]{Only for worker-to-server communication in distributed training, weights are accumulated with float32}

\subsection{Training Curves and Regularization}
\begin{figure}[h]
\begin{center}
\includegraphics[width=5.5in]{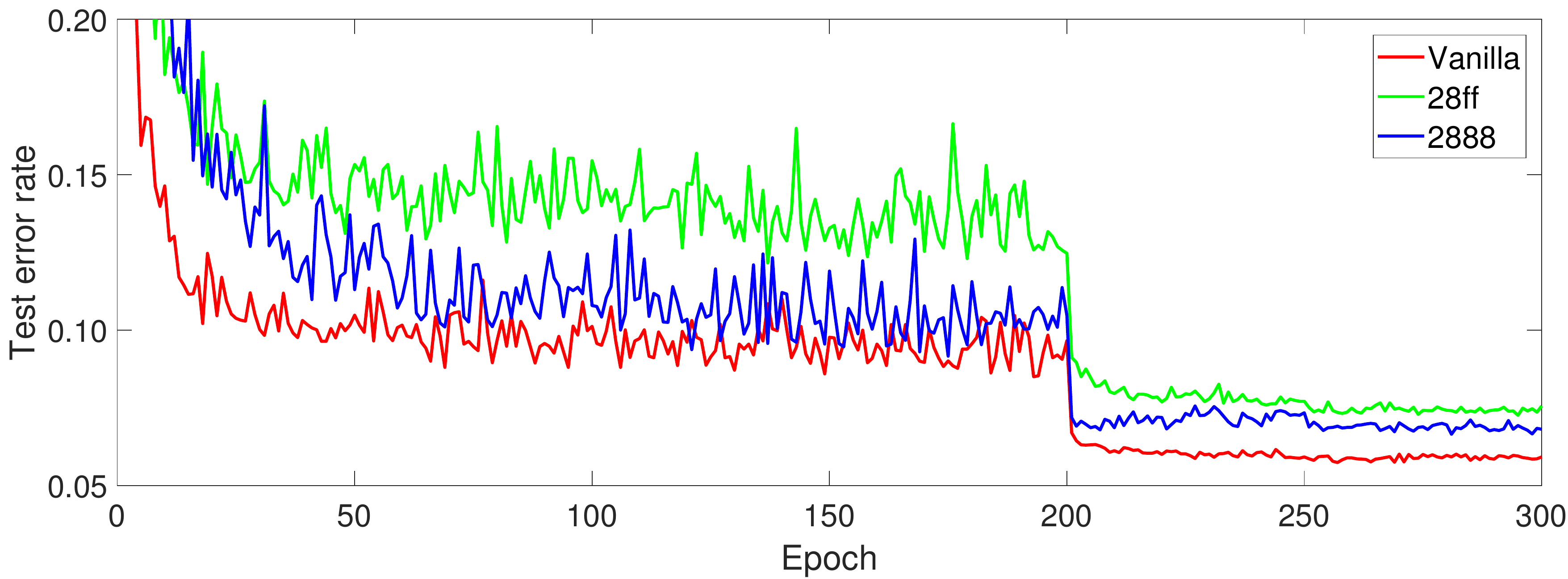}
\end{center}
\caption{Training curves of WAGE variations and a vanilla CNN on CIFAR10.}
\label{fig2}
\end{figure}

We further compare WAGE variations and a vanilla CNN on CIFAR10. The vanilla CNN has the same VGG-like architecture described above except that none quantization of any operand or operation is applied. We add batch normalization in each layer and \emph{Softmax} for the last layer, replace SSE with cross-entropy criterion, and then use a L2 weight decay of 1e-4 and momentum of $0.9$ for training. The learning rate is set to 0.1 and divided by 10 at epoch 200 and epoch 250. For variations of WAGE, pattern 28ff has no quantization nodes in backpropagation. Although the 28ff pattern has the same optimizer and learning rate annealing method as the vanilla pattern, we find that weight updates are decreased by the rescale factor $\alpha$ in Equation \ref{eqn6}. Therefore, the learning rate for 28ff is amplified and tuned, which reduces the error rate by 3\%.
Figure~\ref{fig2} shows the training curves of three counterparts. It can be seen that the 2888 pattern has comparable convergence rate to the vanilla CNN, better accuracy than those who only discretize weights and activations in inference time, though slightly more volatile. The discretization of backpropagation somehow acts as another type of regularization and have significant error rate drop when decreasing learning rate $\eta$.

\subsection{Bitwidth of Errors}
\begin{figure}[h]
\begin{center}
\includegraphics[width=5.5in]{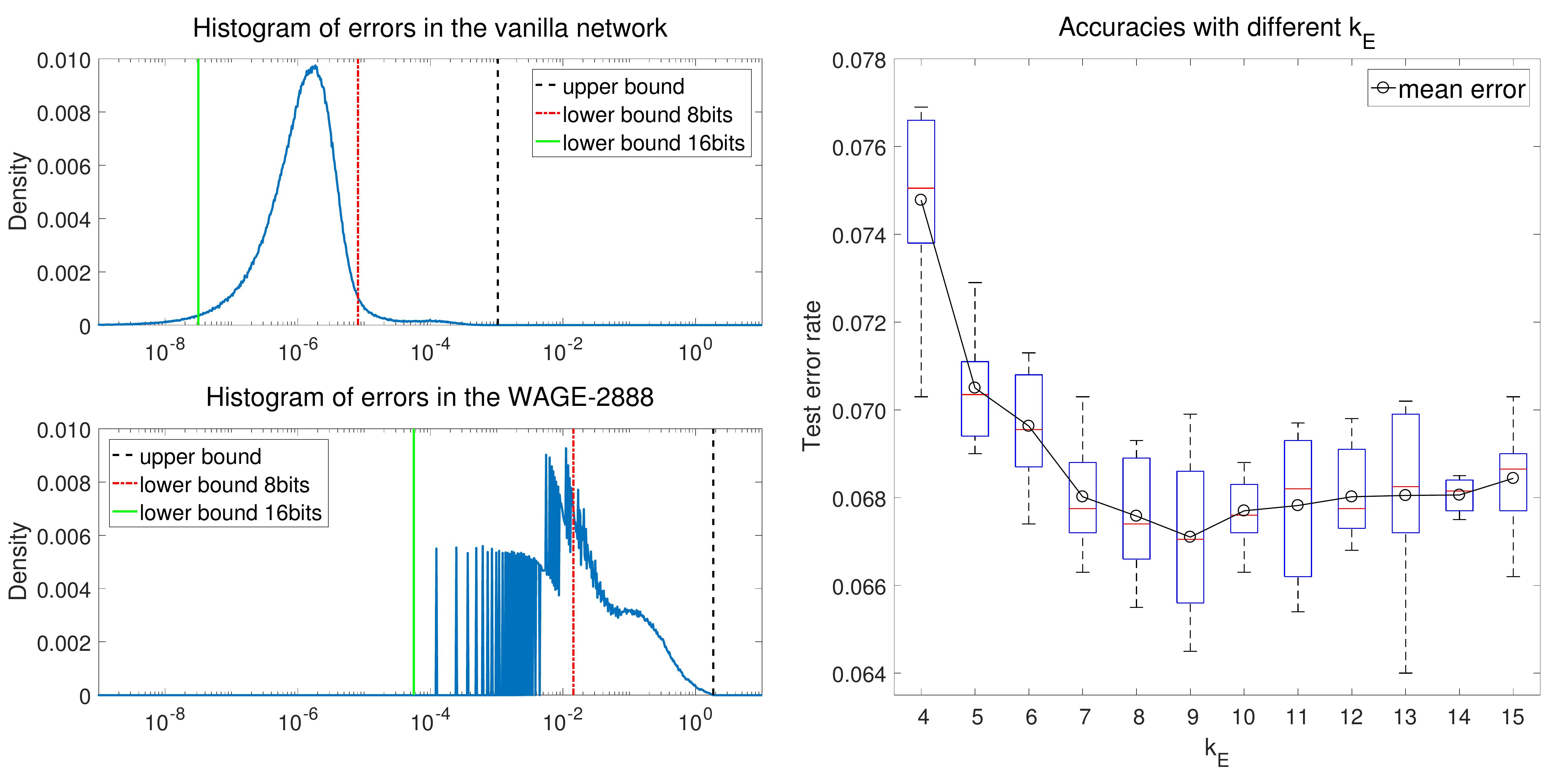}
\end{center}
\caption{Left are histograms of errors $\bm{e}$ for same layer in vanilla network and WAGE-2888 network. Upper boundaries are the $max\{|\bm{e}|\}$ while lower boundaries are determined by the bitwidth $k_E$. The 10 run accuracies of different $k_E$ are shown on the right.}
\label{fig3}
\end{figure}

The bitwidth $k_E$ is set to 8 as default in previous experiments. To further explore a proper bitwidth and its truncated boundary, we firstly export errors from vanilla CNN for CIFAR10 after 100 training epochs. The histogram of errors in the last convolution layer among 128 mini-batch data is shown in Figure~\ref{fig3}. It is obvious that errors approximately obey logarithmic normal distribution where values are relatively small and have significantly large range. When quantized with $k_E$-bit integers, a proper window function should be chosen to truncate the distribution while retaining the approximate orientations for backpropagation. For more details about the layerwise histograms of all W, A, G, E operands, see Figure~\ref{fig5}.

Firstly, the upper (right) boundary is immobilized to the maximum absolute value among all elements in errors as described in Equation~\ref{eqn10}. Then the left boundary will be based on the bitwidth $k_E$. We conduct a series of experiments for $k_E$ ranging from 4 to 15. The boxplot in Figure~\ref{fig3} indicates that 4-8 bits of errors represented by integers are enough for CIFAR10 classification task. Bitwidth 8 is chosen as default to match the 8-bit image color levels and most operands in the micro control unit (MCU). The histogram of errors in the same layer of WAGE-2888 shows that after being shifted and quantized layer by layer, the distribution of errors reshapes and mostly aggregates into truncated window. Thus, most information for orientations is retained. Besides, the smaller values in errors have negligible effects on previous orientations though accumulated layer by layer, which are partially discarded in quantization.

Since the width of the window has been optimized, we left-shift the window with factor $\gamma$ to explore its horizontal position. The right boundary can be formulated as $max\{|\bm{e}|\}/\gamma$. Table~\ref{tab2} shows the effect of shifting errors: although large values are in the minority, they play critical roles for backpropagation training while the majority with small values actually act as noises.


\begin{table}[h]
\caption{Test error rates (\%) on CIFAR10 when left-shift upper boundary with factor $\gamma$.}
\begin{center}
\begin{tabular}{ c|c|c|c|c }
\hline
$\gamma$ & 1 & 2 & 4 & 8 \\
\hline
error & 6.78 & 7.31 & 8.08 & 16.92 \\
\hline
\end{tabular}
\end{center}
\label{tab2}
\end{table}

\subsection{Bitwidth of Gradients}
The bitwidth $k_G$ is set to 8 as default in previous experiments. Although weights are propagated with ternary values in inference and achieve $16\times$ compression rate than float32 weights, they are saved and accumulated in a relatively higher bitwidth (8 bits) for backpropagation training. Therefore, the overall compression rate is only $4\times$. The inconsistent bitwidth between weight updates $k_G$ and their effects in inference $k_W$ provides indispensable buffer space. Otherwise, there might be too many weights changing their ternary values in each iteration, making training very slow and unstable. To further explore a proper bitwidth for gradients, we use WAGE 2-8-8-8 in CIFAR10 as baseline and range $k_G$ from 2 to 12, the learning rate $\eta$ is divided by 2 every time the $k_G$ decreases 1 bit to keep approximately equal weights accumulation in large number of iterations. Results from Table~\ref{tab3} show the effect of $k_G$ and indicate the similar bitwidth requirement as previous experiments for $k_E$.

\begin{table}[h]
\caption{Test error rates (\%) on CIFAR10 with different $k_G$.}
\begin{center}
\begin{tabular}{ c|c|c|c|c|c|c|c|c|c|c|c }
\hline
$k_G$ & 2 & 3 & 4 & 5 & 6 & 7 & 8 & 9 & 10 & 11 & 12  \\
\hline
error & 54.22 & 51.57 & 28.22 & 18.01 & 11.48 & 7.61 & 6.78 & 6.63 & 6.43 & 6.55 & 6.57\\
\hline
\end{tabular}
\end{center}
\label{tab3}
\end{table}

For ImageNet implementation, we conduct six patterns to show bitwidth requirements: 2888 from Table~\ref{tab1}, 288C for more accurate errors (12 bits), 28C8 for larger buffer space, 28f8 for none quantization of gradients, 28ff for errors and gradients in float32 as unlimited case and its BN counterpart. The accuracy of original AlexNet reproduction is reported as baseline. Learning rate $\eta$ is set to 64 and divided by 8 in 28C8 pattern, 0.01 and divided by 10 in 28f8, 28ff counterparts and vanilla AlexNet. We observe overfitting when increasing $k_G$ thus add L2 weight decay of 1e-4, 1e-4 and 5e-4 for 28f8, 28ff and 28ff-BN patterns, respectively. In table~\ref{tab4}, the comparison between pattern 28C8 and 288C reveals that it might be more important to make more buffer space $k_G$ for gradient accumulation than to keep high-resolution orientation $k_E$. Besides, when it comes to ImageNet dataset, the gradient accumulation, i.e., the bit width of gradients ($k_G$) and batch normalization become more important \citep{li2017training} since samples in training set are so variant.

To avoid external memory consumption of full-precision weights during training, \citet{deng2018gxnor} achieved 1-bit weights representation in both training and inference. They use a much larger mini-batch size of 1000 and float32 backpropagation dataflow to accumulate more precise weight updates, equally compensating the buffer space in WAGE provided by external bits of $k_G$. However, large batch size will dramatically increase total training time, counteracting the speed benefits brought by integer arithmetic units. Besides, intermediate variables like feature maps often consume much more memory than weights and linearly correlated with mini-batch size. Therefore, we apply bigger $k_G$ for better convergence rate, accuracy and lower memory usage.

\begin{table}[h]
\caption{Top-5 error rates (\%) on ImageNet with different $k_G$ and $k_E$.}
\begin{center}
\begin{tabular}{ c|c|c|c|c|c|c|c }
\hline
Pattern & Vanilla & 28ff-BN & 28ff & 28f8 & 28C8  & 288C  & 2888  \\
\hline
error   & 19.29   & 20.67   & 24.14 & 23.92 & 26.88 & 28.06 & 27.82 \\
\hline
\end{tabular}
\end{center}
\label{tab4}
\end{table}

\section{Discussion and Future Work}

The goal of this work is to demonstrate potentials of applying training and inference with low-bitwidth integers in DNNs.
Compared with FP16, 8-bit integer operations will not only reduce the energy and area costs for IC design (about $5\times$, see Table~\ref{tab5}), but also halve the memory accesses costs and memory size requirements during training, which will greatly benefit mobile devices with on-site learning capability.
There are some points not involved in this work but yet to be improved or solved in future algorithm developments and hardware deployment.

\textbf{MAC Operation}: WAGE framework is mainly tested with 2-8-8-8 bitwidth configuration, which means that though there are no multiplications during inference with ternary weights, MACs are still needed to calculate $\bm{g}$ in training. Possible solution is 2-2-8-8 pattern if we do not consider the matching of bitwidths between $\bm{a}$ and $\bm{e}$. However, ternary $\bm{a}$ will dramatically slow down convergence and hurt accuracy since $Q(x,2)$ has two relatively high thresholds and clear most outputs of each layer at the beginning of training, this phenomenon is also observed in our BNN replication.

\textbf{Non-linear quantization}: The linear mapping with uniform distance is adopted in WAGE for its simplicity. However, non-linear quantization method like logarithmic representation \citep{miyashita2016convolutional,zhou2017incremental} might be more efficient because the weights and activations in a trained network naturally have logarithmic normal distributions as shown in Figure~\ref{fig3}. Besides, values in logarithmic representation have much larger range with fewer bits than fixed-point representation and are naturally encoded in digital hardware. It is promising to training DNNs with integers encoded with logarithmic representation.

\textbf{Normalization}: Normalization layers like \emph{Softmax} and batch normalization are avoided or removed in some WAGE demonstrations. We think normalizations are essential for end-to-end multi-channel perception where sensors with different modalities have different input distributions, as well as cross-model features encoding and cognition where information from different branches gather to form higher-level representations. Therefore, a better way to quantize normalization is of great interest in further studies.

\begin{table}[h]
\caption{Rough relative costs in 45nm 0.9V from \citet{sze2017efficient}.}
\begin{center}
\begin{tabular}{ c|cc|cc }
\toprule
\multirow{2}{*}{Operation} & \multicolumn{2}{c|}{Energy(pJ)} & \multicolumn{2}{c}{Area(\textmu $\rm m^2$)} \\
\cline{2-5}
 & MUL & ADD & MUL & ADD \\
\hline
8-bit INT & 0.2 pJ  & 0.03 pJ &  282 & 36       \\
16-bit FP     & 1.1 pJ  & 0.40 pJ &  1640 & 1360       \\
32-bit FP     & 3.7 pJ  & 0.90 pJ &   7700 & 4184      \\
\bottomrule
\end{tabular}
\end{center}
\label{tab5}
\end{table}

\section{Conclusion}

WAGE empowers pure low-bitwidth integer dataflow in DNNs for both training and inference. We introduce a new initialization method and a layer-wise constant scaling factor to replace batch normalization, which is a pain spot for network quantization. Many other components for training are also considered or simplified by alternative solutions. In addition, the bitwidth requirements for error computation and gradient accumulation are explored. Experiments reveal that we can quantize relative values of gradients, as well as discard the majority of small values and their orders of magnitude in backpropagation. Although the accumulation for weights updates are indispensable for stable convergence and final accuracy, there still remain works for compression and memory consumption can be further reduced in training. WAGE achieves state-of-art accuracies on multiple datasets with 2-8-8-8 bitwidth configuration. It is promising for incremental works via fine-tuning, more efficient mapping, quantization of batch normalization, etc. Overall, we introduce a framework without floating-point representation and demonstrate the potential to implement both discrete training and inference on integer-based lightweight ASIC or FPGA with on-site learning capability.

\subsubsection*{Acknowledgments}
This work is partially supported by the Project of NSFC (61327902), the SuZhou-Tsinghua innovation leading program (2016SZ0102), the National Natural Science Foundation of China (61603209) and the Independent Research Plan of Tsinghua University (20151080467). We discuss a lot with Peng Jiao and Lei Deng, gratefully acknowledge for their thoughtful comments.

\bibliography{iclr2018_conference}

\begin{thebibliography}{30}
\providecommand{\natexlab}[1]{#1}
\providecommand{\url}[1]{\texttt{#1}}
\expandafter\ifx\csname urlstyle\endcsname\relax
  \providecommand{\doi}[1]{doi: #1}\else
  \providecommand{\doi}{doi: \begingroup \urlstyle{rm}\Url}\fi

\bibitem[Abadi et~al.(2016)Abadi, Barham, Chen, Chen, Davis, Dean, Devin,
  Ghemawat, Irving, Isard, et~al.]{abadi2016tensorflow}
Mart{\'\i}n Abadi, Paul Barham, Jianmin Chen, Zhifeng Chen, Andy Davis, Jeffrey
  Dean, Matthieu Devin, Sanjay Ghemawat, Geoffrey Irving, Michael Isard, et~al.
\newblock Tensorflow: A system for large-scale machine learning.
\newblock In \emph{OSDI}, volume~16, pp.\  265--283, 2016.

\bibitem[Chen et~al.(2017)Chen, Krishna, Emer, and Sze]{chen2017eyeriss}
Yu-Hsin Chen, Tushar Krishna, Joel~S Emer, and Vivienne Sze.
\newblock Eyeriss: An energy-efficient reconfigurable accelerator for deep
  convolutional neural networks.
\newblock \emph{IEEE Journal of Solid-State Circuits}, 52\penalty0
  (1):\penalty0 127--138, 2017.

\bibitem[Courbariaux et~al.(2015)Courbariaux, Bengio, and
  David]{courbariaux2015binaryconnect}
Matthieu Courbariaux, Yoshua Bengio, and Jean-Pierre David.
\newblock Binaryconnect: Training deep neural networks with binary weights
  during propagations.
\newblock In \emph{Advances in Neural Information Processing Systems}, pp.\
  3123--3131, 2015.

\bibitem[Deng et~al.(2018)Deng, Jiao, Pei, Wu, and Li]{deng2018gxnor}
Lei Deng, Peng Jiao, Jing Pei, Zhenzhi Wu, and Guoqi Li.
\newblock Gxnor-net: Training deep neural networks with ternary weights and
  activations without full-precision memory under a unified discretization
  framework.
\newblock \emph{Neural Networks}, 2018.

\bibitem[Han et~al.(2015)Han, Mao, and Dally]{han2015deep}
Song Han, Huizi Mao, and William~J Dally.
\newblock Deep compression: Compressing deep neural networks with pruning,
  trained quantization and huffman coding.
\newblock \emph{arXiv preprint arXiv:1510.00149}, 2015.

\bibitem[He et~al.(2015)He, Zhang, Ren, and Sun]{he2015delving}
Kaiming He, Xiangyu Zhang, Shaoqing Ren, and Jian Sun.
\newblock Delving deep into rectifiers: Surpassing human-level performance on
  imagenet classification.
\newblock In \emph{Proceedings of the IEEE international conference on computer
  vision}, pp.\  1026--1034, 2015.

\bibitem[Hinton et~al.(2012)Hinton, Deng, Yu, Dahl, Mohamed, Jaitly, Senior,
  Vanhoucke, Nguyen, Sainath, et~al.]{hinton2012deep}
Geoffrey Hinton, Li~Deng, Dong Yu, George~E Dahl, Abdel-rahman Mohamed, Navdeep
  Jaitly, Andrew Senior, Vincent Vanhoucke, Patrick Nguyen, Tara~N Sainath,
  et~al.
\newblock Deep neural networks for acoustic modeling in speech recognition: The
  shared views of four research groups.
\newblock \emph{IEEE Signal Processing Magazine}, 29\penalty0 (6):\penalty0
  82--97, 2012.

\bibitem[Howard et~al.(2017)Howard, Zhu, Chen, Kalenichenko, Wang, Weyand,
  Andreetto, and Adam]{howard2017mobilenets}
Andrew~G Howard, Menglong Zhu, Bo~Chen, Dmitry Kalenichenko, Weijun Wang,
  Tobias Weyand, Marco Andreetto, and Hartwig Adam.
\newblock Mobilenets: Efficient convolutional neural networks for mobile vision
  applications.
\newblock \emph{arXiv preprint arXiv:1704.04861}, 2017.

\bibitem[Hubara et~al.(2016)Hubara, Courbariaux, Soudry, El-Yaniv, and
  Bengio]{hubara2016binarized}
Itay Hubara, Matthieu Courbariaux, Daniel Soudry, Ran El-Yaniv, and Yoshua
  Bengio.
\newblock Binarized neural networks.
\newblock In \emph{Advances in Neural Information Processing Systems}, pp.\
  4107--4115, 2016.

\bibitem[Ioffe \& Szegedy(2015)Ioffe and Szegedy]{ioffe2015batch}
Sergey Ioffe and Christian Szegedy.
\newblock Batch normalization: Accelerating deep network training by reducing
  internal covariate shift.
\newblock In \emph{International Conference on Machine Learning}, pp.\
  448--456, 2015.

\bibitem[Jouppi et~al.(2017)Jouppi, Young, Patil, Patterson, Agrawal, Bajwa,
  Bates, Bhatia, Boden, Borchers, et~al.]{jouppi2017datacenter}
Norman~P Jouppi, Cliff Young, Nishant Patil, David Patterson, Gaurav Agrawal,
  Raminder Bajwa, Sarah Bates, Suresh Bhatia, Nan Boden, Al~Borchers, et~al.
\newblock In-datacenter performance analysis of a tensor processing unit.
\newblock In \emph{Proceedings of the 44th Annual International Symposium on
  Computer Architecture}, pp.\  1--12. ACM, 2017.

\bibitem[Krizhevsky et~al.(2012)Krizhevsky, Sutskever, and
  Hinton]{krizhevsky2012imagenet}
Alex Krizhevsky, Ilya Sutskever, and Geoffrey~E Hinton.
\newblock Imagenet classification with deep convolutional neural networks.
\newblock In \emph{Advances in neural information processing systems}, pp.\
  1097--1105, 2012.

\bibitem[LeCun et~al.(1998)LeCun, Bottou, Bengio, and
  Haffner]{lecun1998gradient}
Yann LeCun, L{\'e}on Bottou, Yoshua Bengio, and Patrick Haffner.
\newblock Gradient-based learning applied to document recognition.
\newblock \emph{Proceedings of the IEEE}, 86\penalty0 (11):\penalty0
  2278--2324, 1998.

\bibitem[Lee et~al.(2015)Lee, Xie, Gallagher, Zhang, and Tu]{lee2015deeply}
Chen-Yu Lee, Saining Xie, Patrick Gallagher, Zhengyou Zhang, and Zhuowen Tu.
\newblock Deeply-supervised nets.
\newblock In \emph{Artificial Intelligence and Statistics}, pp.\  562--570,
  2015.

\bibitem[Li et~al.(2016)Li, Zhang, and Liu]{li2016ternary}
Fengfu Li, Bo~Zhang, and Bin Liu.
\newblock Ternary weight networks.
\newblock \emph{arXiv preprint arXiv:1605.04711}, 2016.

\bibitem[Li et~al.(2017)Li, De, Xu, Studer, Samet, and
  Goldstein]{li2017training}
Hao Li, Soham De, Zheng Xu, Christoph Studer, Hanan Samet, and Tom Goldstein.
\newblock Training quantized nets: A deeper understanding.
\newblock In \emph{Advances in Neural Information Processing Systems}, pp.\
  5813--5823, 2017.

\bibitem[Miyashita et~al.(2016)Miyashita, Lee, and
  Murmann]{miyashita2016convolutional}
Daisuke Miyashita, Edward~H Lee, and Boris Murmann.
\newblock Convolutional neural networks using logarithmic data representation.
\newblock \emph{arXiv preprint arXiv:1603.01025}, 2016.

\bibitem[Rastegari et~al.(2016)Rastegari, Ordonez, Redmon, and
  Farhadi]{rastegari2016xnor}
Mohammad Rastegari, Vicente Ordonez, Joseph Redmon, and Ali Farhadi.
\newblock Xnor-net: Imagenet classification using binary convolutional neural
  networks.
\newblock In \emph{European Conference on Computer Vision}, pp.\  525--542.
  Springer, 2016.

\bibitem[Russakovsky et~al.(2015)Russakovsky, Deng, Su, Krause, Satheesh, Ma,
  Huang, Karpathy, Khosla, Bernstein, et~al.]{russakovsky2015imagenet}
Olga Russakovsky, Jia Deng, Hao Su, Jonathan Krause, Sanjeev Satheesh, Sean Ma,
  Zhiheng Huang, Andrej Karpathy, Aditya Khosla, Michael Bernstein, et~al.
\newblock Imagenet large scale visual recognition challenge.
\newblock \emph{International Journal of Computer Vision}, 115\penalty0
  (3):\penalty0 211--252, 2015.

\bibitem[Sheridan et~al.(2017)Sheridan, Cai, Du, Ma, Zhang, and
  Lu]{sheridan2017sparse}
Patrick~M Sheridan, Fuxi Cai, Chao Du, Wen Ma, Zhengya Zhang, and Wei~D Lu.
\newblock Sparse coding with memristor networks.
\newblock \emph{Nature nanotechnology}, 12\penalty0 (8):\penalty0 784, 2017.

\bibitem[Shi et~al.(2015)Shi, Pei, Deng, Wang, Deng, Wang, Zhang, Chen, Zhao,
  Song, et~al.]{shi2015development}
Luping Shi, Jing Pei, Ning Deng, Dong Wang, Lei Deng, Yu~Wang, Youhui Zhang,
  Feng Chen, Mingguo Zhao, Sen Song, et~al.
\newblock Development of a neuromorphic computing system.
\newblock In \emph{Electron Devices Meeting (IEDM), 2015 IEEE International},
  pp.\  4--3. IEEE, 2015.

\bibitem[Silver et~al.(2016)Silver, Huang, Maddison, Guez, Sifre, Van
  Den~Driessche, Schrittwieser, Antonoglou, Panneershelvam, Lanctot,
  et~al.]{silver2016mastering}
David Silver, Aja Huang, Chris~J Maddison, Arthur Guez, Laurent Sifre, George
  Van Den~Driessche, Julian Schrittwieser, Ioannis Antonoglou, Veda
  Panneershelvam, Marc Lanctot, et~al.
\newblock Mastering the game of go with deep neural networks and tree search.
\newblock \emph{Nature}, 529\penalty0 (7587):\penalty0 484--489, 2016.

\bibitem[Simonyan \& Zisserman(2014)Simonyan and Zisserman]{simonyan2014very}
Karen Simonyan and Andrew Zisserman.
\newblock Very deep convolutional networks for large-scale image recognition.
\newblock \emph{arXiv preprint arXiv:1409.1556}, 2014.

\bibitem[Sze et~al.(2017)Sze, Chen, Yang, and Emer]{sze2017efficient}
Vivienne Sze, Yu-Hsin Chen, Tien-Ju Yang, and Joel~S Emer.
\newblock Efficient processing of deep neural networks: A tutorial and survey.
\newblock \emph{Proceedings of the IEEE}, 105\penalty0 (12):\penalty0
  2295--2329, 2017.

\bibitem[Tang et~al.(2017)Tang, Hua, and Wang]{tang2017train}
Wei Tang, Gang Hua, and Liang Wang.
\newblock How to train a compact binary neural network with high accuracy?
\newblock In \emph{Thirty-First AAAI Conference on Artificial Intelligence},
  2017.

\bibitem[Wen et~al.(2017)Wen, Xu, Yan, Wu, Wang, Chen, and Li]{wen2017terngrad}
Wei Wen, Cong Xu, Feng Yan, Chunpeng Wu, Yandan Wang, Yiran Chen, and Hai Li.
\newblock Terngrad: Ternary gradients to reduce communication in distributed
  deep learning.
\newblock In \emph{Advances in Neural Information Processing Systems}, pp.\
  1508--1518, 2017.

\bibitem[Zhang et~al.(2016)Zhang, Bengio, Hardt, Recht, and
  Vinyals]{zhang2016understanding}
Chiyuan Zhang, Samy Bengio, Moritz Hardt, Benjamin Recht, and Oriol Vinyals.
\newblock Understanding deep learning requires rethinking generalization.
\newblock \emph{arXiv preprint arXiv:1611.03530}, 2016.

\bibitem[Zhou et~al.(2017)Zhou, Yao, Guo, Xu, and Chen]{zhou2017incremental}
Aojun Zhou, Anbang Yao, Yiwen Guo, Lin Xu, and Yurong Chen.
\newblock Incremental network quantization: Towards lossless cnns with
  low-precision weights.
\newblock \emph{arXiv preprint arXiv:1702.03044}, 2017.

\bibitem[Zhou et~al.(2016)Zhou, Wu, Ni, Zhou, Wen, and Zou]{zhou2016dorefa}
Shuchang Zhou, Yuxin Wu, Zekun Ni, Xinyu Zhou, He~Wen, and Yuheng Zou.
\newblock Dorefa-net: Training low bitwidth convolutional neural networks with
  low bitwidth gradients.
\newblock \emph{arXiv preprint arXiv:1606.06160}, 2016.

\bibitem[Zhu et~al.(2016)Zhu, Han, Mao, and Dally]{zhu2016trained}
Chenzhuo Zhu, Song Han, Huizi Mao, and William~J Dally.
\newblock Trained ternary quantization.
\newblock \emph{arXiv preprint arXiv:1612.01064}, 2016.

\end{thebibliography}
\bibliographystyle{iclr2018_conference}

\newpage
\appendix
  \renewcommand{\appendixname}{Appendix~\Alph{section}}
\section{Algorithm}
We assume that network structures are defined and initialized with Equation~\ref{eqn5}. The annotations after pseudo code are potential corresponding operations for implementation in a fixed-point dataflow.
\begin{algorithm}
  \caption{Training an $I$-layer net with WAGE method on floating-point-based or integer-based device. Weights, activations, gradients and errors are quantized according to Equations~\ref{eqn8}~-~\ref{eqn13}.}
    \begin{algorithmic}[1]

      \REQUIRE a mini-batch of inputs and targets $(\bm{a}_q^0,\bm{a}^*)$ which are quantized to $k_A$-bit integers, shift-based $\alpha$ for each layer, learning rate scheduler $\eta$, previous weight $\bm{W}$ saved in $k_G$ bits.
      \ENSURE updated weights $\bm{W}_{t+1}$

      \textbf{1. Forward propagation:}\\
      \FOR{$i=1$ to $I$}
        \STATE $\bm{W}_q^i \leftarrow Q_W(\bm{W}^i)$
        \hspace{5.00cm} \#Clip \\
        \STATE $\bm{a}^i \leftarrow ReLU(\bm{a}_q^{i-1} \bm{W}_q^i)$
        \hspace{4.15cm} \#MAC, Clip\\
        \STATE $\bm{a}_q^i \leftarrow Q_A(\bm{a}^i)$
        \hspace{5.48cm} \#Shift, Clip \\
      \ENDFOR \\
      \textbf{2. Back propagation:}\\
      Compute $\bm{e}^I \leftarrow \frac{ \partial{\mathcal{L}} }{ \partial{\bm{a}^I} }$ knowing $\bm{a}^I$ and $\bm{a}^*$
      \hspace{2.18cm} \#Substrate
      \FOR{$i=I$ to $1$}
        \STATE $\bm{e}_q^i \leftarrow Q_E(\bm{e}^i)$
        \hspace{5.52cm} \#Max, Shift, Clip \\
        \STATE $\bm{e}^{i-1} \leftarrow \bm{e}_q^i {\bm{W}_q^i}$
        \hspace{5.40cm} \#MAC, Clip\\
        \STATE $\bm{g}^i \leftarrow {\bm{e}_q^i}^{\mathrm{T}} \bm{a}_q^{i-1}$
        \hspace{5.40cm} \#MAC, Clip\\
        \STATE $\Delta{\bm{W}^i} \leftarrow Q_G(\bm{g}^i)$
        \hspace{5.05cm} \#Max, Shift, Random, Clip\\
        \STATE Update and Clip $\bm{W}^i$ according to Equation~\ref{eqn13}
        \hspace{1cm} \#Update, Clip \\

      \ENDFOR \\

    \end{algorithmic}
\label{algorithm1}
\end{algorithm}

\newpage
\section{Layerwise Histogram}

  \begin{figure}[h]
  \includegraphics[width=5.52in]{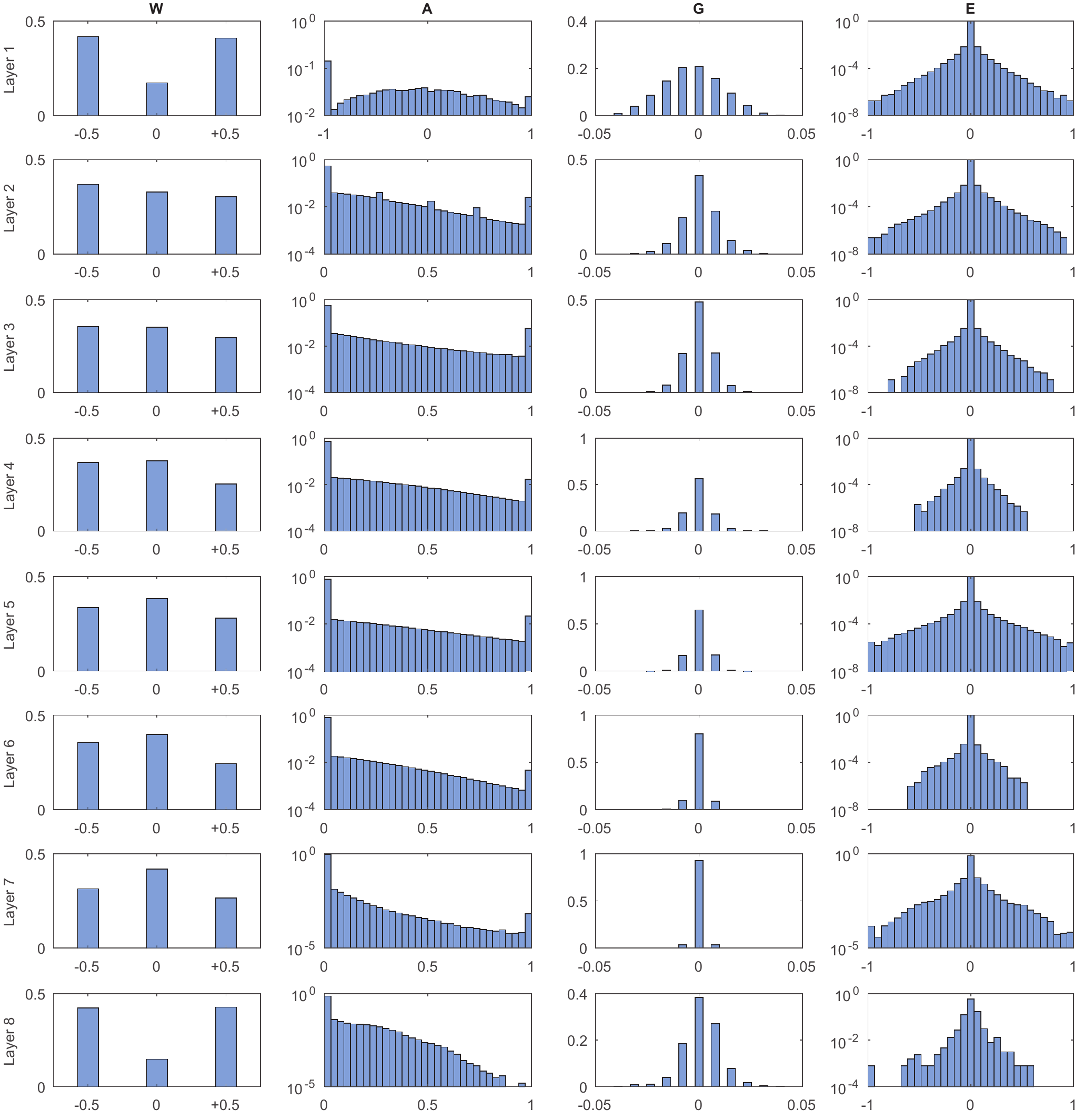}
  \caption{
  Layerwise histograms of a trained VGG-like network with bitwidth configuration: 2-8-8-8 and learning rate $\eta$ equals to 8. The Y-axis represents for probability in W-plots and G-plots, and logarithmic probability in A-plots and E-plots, respectively. In A-plots histograms are one-layer ahead so the first figure shows the quantized input data.}
  \label{fig5}
  \end{figure}

\end{document}